# Extension of Inagaki General Weighted Operators

# and

# A New Fusion Rule Class of Proportional Redistribution of Intersection Masses

Florentin Smarandache
Chair of Math & Sciences Depart.
University of New Mexico, Gallup, USA

**Abstract**.
In this paper we extend Inagaki Weighted Operators fusion rule (WO) [see 1, 2] in information fusion by doing redistribution of not only the conflicting mass, but also of masses of non-empty intersections, that we call <u>Double Weighted Operators</u> (DWO).
Then we propose a new fusion rule <u>Class of Proportional Redistribution of Intersection Masses</u> (CPRIM), which generates many interesting particular fusion rules in information fusion.
Both formulas are presented for 2 and for n ≥ 3 sources.
An application and comparison with other fusion rules are given in the last section.

**1. Introduction**.

Let $\theta = \{\theta_1, \theta_2, ..., \theta_n\}$, for $n \geq 2$, be the frame of discernment, and $S^\theta = (\theta, \cup, \cap, \tau)$ its super-power set, where τ(x) means complement of x with respect to the total ignorance.

Let $I_t$ = total ignorance = θ₁∪ θ₂∪…∪θₙ, and Φ be the empty set.

$S^\theta = 2\,^\wedge\theta_{\text{refined}} = 2^\wedge(2^\wedge\theta) = D^{\theta \cup \theta c}$, when refinement is possible, where θ_c = {τ(θ₁), τ(θ₂), …, τ(θₙ)}.

We consider the general case when the domain is $S^\theta$, but $S^\theta$ can be replaced by $D^\theta = (\theta, \cup, \cap)$ or by $2^\theta = (\theta, \cup)$ in all formulas from below.

Let $m_1(\cdot)$ and $m_2(\cdot)$ be two normalized masses defined from $S^\theta$ to $[0,1]$.

We use the conjunction rule to first combine $m_1(\cdot)$ with $m_2(\cdot)$ and then we redistribute the mass of $m(X \cap Y) \neq 0$, when $X \cap Y = \Phi$.

Let's denote $m_{2\cap}(A) = (m_1 \oplus m_2)(A) = \sum_{\substack{X,Y \in S^\theta \\ (X \cap Y)=A}} m_1(X)m_2(Y)$ using the conjunction rule.

Let's note the set of intersections by:

$$S_\cap = \left\{ \begin{array}{l} X \in S^\theta \mid X = y \cap z, \text{ where } y, z \in S^\theta \setminus \{\Phi\}, \\ X \text{ is in a canonical form, and} \\ X \text{ contains at least an } \cap \text{ symbol in its formula} \end{array} \right\}. \quad (1)$$

In conclusion, $S_\cap$ is a set of formulas formed with singletons (elements from the frame of discernment), such that each formula contains at least an intersection symbol $\cap$, and each formula is in a canonical form (easiest form).

For example: $A \cap A \notin S_\cap$ since $A \cap A$ is not a canonical form, and $A \cap A = A$. Also, $(A \cap B) \cap B$ is not in a canonical form but $(A \cap B) \cap B = A \cap B \in S_\cap$.

Let

$S_\cap^\Phi$ = the set of all empty intersections from $S_\cap$,

and

$S_{\cap,r}^{non\Phi}$ = {the set of all non-empty intersections from $S_\cap^{non\Phi}$ whose masses are redistributed to other sets, which actually depends on the sub-model of each application}.

## 2. Extension of Inagaki General Weighted Operators (WO).

Inagaki general weighted operator $(WO)$ is defined for two sources as:
$$\forall A \in 2^\theta \setminus \{\Phi\}, \quad m_{(WO)}(A) = \sum_{\substack{X,Y \in 2^\theta \\ (X \cap Y)=A}} m_1(X)m_2(Y) + W_m(A) \cdot m_{2\cap}(\Phi), \quad (2)$$

where
$$\sum_{X \in 2^\theta} W_m(X) = 1 \text{ and all } W_m(\cdot) \in [0,1]. \quad (3)$$

So, the conflicting mass is redistributed to non-empty sets according to these weights $W_m(\cdot)$.

In the extension of this $WO$, which we call the Double Weighted Operator $(DWO)$, we redistribute not only the conflicting mass $m_{2\cap}(\Phi)$ but also the mass of some (or all) non-empty intersections, i.e. those from the set $S_{\cap,r}^{non\Phi}$, to non-empty sets from $S^\theta$ according to some weights $W_m(\cdot)$ for the conflicting mass (as in WO), and respectively according to the weights $V_m(\cdot)$ for the non-conflicting mass of the elements from the set $S_{\cap,r}^{non\Phi}$:

$$\forall A \in (S^\theta \setminus S_{\cap,r}^{non\Phi}) \setminus \{\Phi\}, \quad m_{DWO}(A) = \sum_{\substack{X,Y \in S^\theta \\ (X \cap Y)=A}} m_1(X)m_2(Y) + W_m(A) \cdot m_{2\cap}(\Phi) + V_m(A) \cdot \sum_{z \in S_{\cap,r}^{non\Phi}} m_{2\cap}(z),$$

(4)

where
$$\sum_{X \in S^\theta} W_m(X) = 1 \text{ and all } W_m(\cdot) \in [0,1], \text{ as in (3)}$$

and
$$\sum_{z \in S_{\cap,r}^{non\Phi}} V_m(z) = 1 \text{ and all } V_m(\cdot) \in [0,1]. \tag{5}$$

In the free and hybrid modes, if no non-empty intersection is redistributed, i.e. $S_{\cap,r}^{non\Phi}$ contains no elements, $DWO$ coincides with $WO$.

In the Shafer's model, always $DWO$ coincides with $WO$.

For $s \geq 2$ sources, we have a similar formula:

$$\forall A \in \left(S^\theta \setminus S_{\cap,r}^{non\Phi}\right) \setminus \{\Phi\}, \ m_{DWO}(A) = \sum_{\substack{X_1, X_2, \ldots, X_n \in S^\theta \\ \bigcap_{i=1}^s X_i = A}} \prod_{i=1}^s m_i(X_i) + W_m(A) \cdot m_{s\cap}(\Phi) + V_m(A) \cdot \sum_{z \in S_{\cap,r}^{non\Phi}} m_{s\cap}(z)$$

(6)

with the same restrictions on $W_m(\cdot)$ and $V_m(\cdot)$.

## 3. A Fusion Rule Class of Proportional Redistribution of Intersection Masses

For $A \in \left(S^\theta \setminus S_{\cap,r}^{non\Phi}\right) \setminus \{\Phi, I_t\}$ for two sources we have:

$$m_{CPRIM}(A) = m_{2\cap}(A) + f(A) \cdot \sum_{\substack{X,Y \in S^\theta \\ \{\Phi = X \cap Y \text{ and } A \subseteq M\} \\ \text{or } \{\Phi \neq X \cap Y \in S_{\cap,r}^{non\Phi} \text{ and } A \subseteq N\}}} \frac{m_1(X)m_2(Y)}{\sum_{z \subseteq M} f(z)}, \tag{7}$$

where $f(X)$ is a function directly proportional to $X$, $f : S^\theta \to [0, \infty]$. (8)

For example, $f(X) = m_{2\cap}(X)$, or (9)

$f(X) = card(X)$, or

$f(X) = \dfrac{card(X)}{card(M)}$ (ratio of cardinals), or

$f(X) = m_{2\cap}(X) + card(X)$, etc.;

and $M$ is a subset of $S^\theta$, for example: (10)

$M = \tau(X \cup Y)$, or

$M = (X \cup Y)$, or

$M$ is a subset of $X \cup Y$, etc.,

where $N$ is a subset of $S^\theta$, for example: (11)

$N = X \cup Y$, or

$N$ is a subset of $X \cup Y$, etc.

And

$$m_{CPRIM}(I_t) = m_{2\cap}(I_t) + \sum_{\substack{X,Y \in S^\theta \\ \{X \cap Y = \Phi \text{ and } (M = \Phi \text{ or } \sum_{z \subseteq M} f(z) = 0)\}}} m_1(X) m_2(Y). \qquad (12)$$

These formulas are easily extended for any $s \geq 2$ sources $m_1(\cdot), m_2(\cdot), ..., m_s(\cdot)$.
Let's denote, using the conjunctive rule:

$$m_{s\cap}(A) = (m_1 \oplus m_2 \oplus ... \oplus m_s)(A) = \sum_{\substack{X_1, X_2, ..., X_s \in S^{\wedge\Theta} \\ \bigcap_{i=1}^{s} X_i = A}} \prod_{i=1}^{s} m_i(x_i) \qquad (13)$$

$$m_{CPRIM}(A) = m_{s\cap}(A) + f(A) \cdot \sum_{\substack{X_1, X_2, ..., X_n \in S^\theta \\ \{\Phi = \bigcap_{i=1}^{s} X_i \text{ and } A \subseteq M\} \\ \text{or } \{\Phi \neq \bigcap_{i=1}^{s} X_i \in S_{\cap,r}^{non\Phi} \text{ and } A \subseteq N\}}} \frac{\prod_{i=1}^{s} m_i(X_i)}{\sum_{z \subseteq M} f(z) \neq 0} \qquad (14)$$

where $f(\cdot)$, $M$, and $N$ are similar to the above where instead of $X \cup Y$ (for two sources) we take $X_1 \cup X_2 \cup ... \cup X_s$ (for s sources), and instead of $m_{2\cap}(X)$ for two sources we take $m_{s\cap}(X)$ for $s$ sources.

**4. Application and Comparison with other Fusion Rules.**
Let's consider the frame of discernment $\Theta = \{A, B, C\}$, and two independent sources $m_1(.)$ and $m_2(.)$ that provide the following masses:

|        | A   | B   | C   | A∪B∪C |
|--------|-----|-----|-----|-------|
| $m_1(.)$ | 0.3 | 0.4 | 0.2 | 0.1   |
| $m_2(.)$ | 0.5 | 0.2 | 0.1 | 0.2   |

Now, we apply the conjunctive rule and we get:

|           | A    | B    | C    | A∪B∪C | A∩B  | A∩C  | B∩C  |
|-----------|------|------|------|-------|------|------|------|
| $m_{12\cap}(.)$ | 0.26 | 0.18 | 0.07 | 0.02  | 0.26 | 0.13 | 0.08 |

Suppose that all intersections are non-empty {this case is called: free DSm (Dezert-Smarandache) Model}. See below the Venn Diagram using the Smarandache codification [3]:

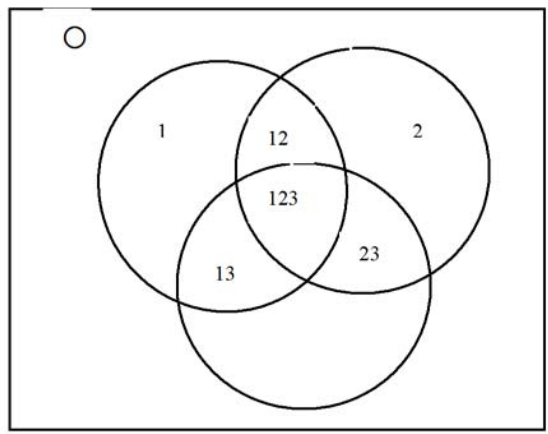

Applying DSm Classic rule, which is a generalization of classical conjunctive rule from the fusion space $(\Theta, \cup)$, called *power set*, when all hypotheses are supposed exclusive (i.e. all intersections are empty) to the fusion space $(\Theta, \cup, \cap)$, called *hyper-power set*, where hypotheses are not necessarily exclusive (i.e. there exist non-empty intersections), we just get:

|  | A | B | C | A∪B∪C | A∩B | A∩C | B∩C |
|---|---|---|---|---|---|---|---|
| $m_{DSmC}(.)$ | 0.26 | 0.18 | 0.07 | 0.02 | 0.26 | 0.13 | 0.08 |

DSmC and the Conjunctive Rule have the same formula, but they work on different fusion spaces.

Inagaki rule was defined on the fusion space $(\Theta, \cup)$. In this case, since all intersections are empty, the total conflicting mass, which is $m_{12\cap}(A\cap B) + m_{12\cap}(A\cap C) + m_{12\cap}(B\cap C) = 0.26 + 0.13 + 0.08 = 0.47$, and this is redistributed to the masses of A, B, C, and A∪B∪C according to some weights $w_1$, $w_2$, $w_3$, and $w_4$ respectively, depending to each particular rule, where: $0 \le w_1, w_2, w_3, w_4 \le 1$ and $w_1 + w_2 + w_3 + w_4 = 1$. Hence

|  | A | B | C | A∪B∪C |
|---|---|---|---|---|
| $m_{Inagaki}(.)$ | $0.26+0.47w_1$ | $0.18+0.47w_2$ | $0.07+0.47w_3$ | $0.02+0.47w_4$ |

Yet, Inagaki rule can also be straightly extended from the power set to the hyper-power set.

Suppose in DWO the user finds out that the hypothesis B∩C is not plausible, therefore $m_{12\cap}(B\cap C) = 0.08$ has to be transferred to the other non-empty elements: A, B, C, A∪B∪C, A∩B, A∩C, according to some weights $v_1$, $v_2$, $v_3$, $v_4$, $v_5$, and $v_6$ respectively, depending to the particular version of this rule is chosen, where:
$0 \le v_1, v_2, v_3, v_4, v_5, v_6 \le 1$ and $v_1 + v_2 + v_3 + v_4 + v_5 + v_6 = 1$. Hence

|  | A | B | C | A∪B∪C | A∩B | A∩C |
|---|---|---|---|---|---|---|
| $m_{DWO}(.)$ | $0.26+0.08v_1$ | $0.18+0.08v_2$ | $0.07+0.08v_3$ | $0.02+0.08v_4$ | $0.26+0.08v_5$ | $0.13+0.08v_6$ |

Now, since CPRIM is a particular case of DWO, but CPRIM is a class of fusion rules, let's consider a sub-particular case for example when the redistribution of $m_{12\cap}(B\cap C) = 0.08$ is done proportionally with respect to the DSm cardinals of B and C which are both equal to 4. DSm cardinal of a set is equal to the number of disjoint parts included in that set upon the Venn Diagram (see it above).
Therefore 0.08 is split equally between B and C, and we get:

|   | A | B | C | A∪B∪C | A∩B | A∩C |
|---|---|---|---|---|---|---|
| $m_{CPRIMcard}(.)$ | 0.26 | 0.18+0.04=0.22 | 0.07+0.04=0.11 | 0.02 | 0.26 | 0.13 |

Applying one or another fusion rule is still debating today, and this depends on the hypotheses, on the sources, and on other information we receive.

**5. Conclusion**.
A generalization of Inagaki rule has been proposed in this paper, and also a new class of fusion rules, called **Class of Proportional Redistribution of Intersection Masses (CPRIM)**, which generates many interesting particular fusion rules in information fusion.

**References**:

[1]  T. Inagaki, Independence Between Safety-Control Policy and Multiple-Sensors Schemes via Dempster-Shafer Theory, IEEE Transaction on Reliability, 40, 182-188, 1991.

[2]  E. Lefèbvre, O. Colot, P. Vannoorenberghe, Belief Function Combination and Conflict Management, Information Fusion 3, 149-162, 2002.

[3] F. Smarandache, J. Dezert (editors), Advances and Applications of DSmT for Information Fusion, Collective Works, Vol. 2, Am. Res. Press, 2004.

June 2008